\documentclass[11pt,a4paper]{article}
\usepackage[hyperref]{emnlp-ijcnlp-2019} 
\usepackage{algorithm}
\usepackage{color}
\usepackage{graphicx}
\usepackage{subfigure}
\usepackage{url}
\usepackage[noend]{algpseudocode}
\usepackage{times}
\usepackage{latexsym}
\usepackage{url}
\usepackage{times}
\usepackage{booktabs}
\usepackage{verbatim}
\usepackage{helvet}  
\usepackage{courier}  
\usepackage{url}  
\usepackage{graphicx}  
\usepackage{booktabs}
\usepackage{amsmath}
\usepackage{comment}
\usepackage{amsfonts}
\usepackage{xcolor}
\usepackage{multirow}
\usepackage{verbatim}
\usepackage{multicol}
\usepackage{subfigure}
\usepackage{url}
\usepackage{caption}
\usepackage{array}
\usepackage{color}

\aclfinalcopy 

\title{Transfer Learning for Relation Extraction via  Relation-Gated  Adversarial  Learning}

\author{
Ningyu Zhang\textsuperscript{1,2,3}   \quad
Shumin Deng\textsuperscript{1,3}    \quad
 Zhanlin Sun\textsuperscript{1,3}  \quad
  Jiaoyan  Chen\textsuperscript{4}  \quad
  \textbf{Wei Zhang\textsuperscript{2,3}} \quad
  Huajun Chen\textsuperscript{1,3}\thanks{\quad Corresponding author.} \\
 1. College of Computer Science and Technology, Zhejiang University\\
 2. Alibaba Group \\
 3. AZFT\thanks{Alibaba-Zhejiang University Frontier Technology Research Center}  Joint Lab for Knowledge Engine\\
 4. Department of Computer Science, Oxford University\\
  \tt \{3150105645,231sm,huajunsir\}@zju.edu.cn \\
  \tt jiaoyan.chen@cs.ox.ac.uk,\{ningyu.zny,lantu.zw\}@alibaba-inc.com}
\begin{document}
\maketitle
\begin{abstract}
Relation extraction aims to extract relational facts from sentences. Previous models mainly rely on manually labeled datasets, seed instances or human-crafted patterns, and distant supervision. However, the human annotation is expensive, while human-crafted patterns suffer from semantic drift and distant supervision samples are usually noisy. Domain adaptation methods enable leveraging labeled data from a different but related domain. However, different domains usually have various textual relation descriptions and different label space (the source label space is usually a  superset of the target label space).  To solve these problems,  we propose a novel model of  relation-gated adversarial learning   for relation extraction, which extends the adversarial based domain adaptation.  Experimental results have shown that the proposed approach outperforms previous domain adaptation methods regarding partial domain adaptation and can improve the accuracy of distance supervised relation extraction through fine-tuning.
\end{abstract}

\section{Introduction}

Relation extraction (RE) is devoted to extracting relational facts from sentences, which can be applied to many natural language processing (NLP) applications such as knowledge base construction  \cite{wu2010open}  and question answering \cite{dai2016cfo}. Given a sentence with an entity pair ($e_1$,$e_2$), this task aims to identify the relation between $e_1$ and $e_2$.

Typically, existing methods follow the supervised learning paradigm, and they require extensive annotations from domain experts, which are expensive and time-consuming. To alleviate such drawbacks, bootstrap learning has been proposed to build relation extractors with a small set of seed instances or human-crafted patterns, but it suffers from the semantic drift problem \cite{nakashole2011scalable}. Besides, distant supervision (DS) methods leverage existing relational pairs of Knowledge Graphs (KGs) such as Freebase to automatically generate training data \cite{mintz2009distant}. However, because of the incompleteness of KGs and a large number of relations among entities, generating sufficient noise-free labels via DS is still prohibitive.
\begin{figure}
\centering
\includegraphics [width=0.48\textwidth]{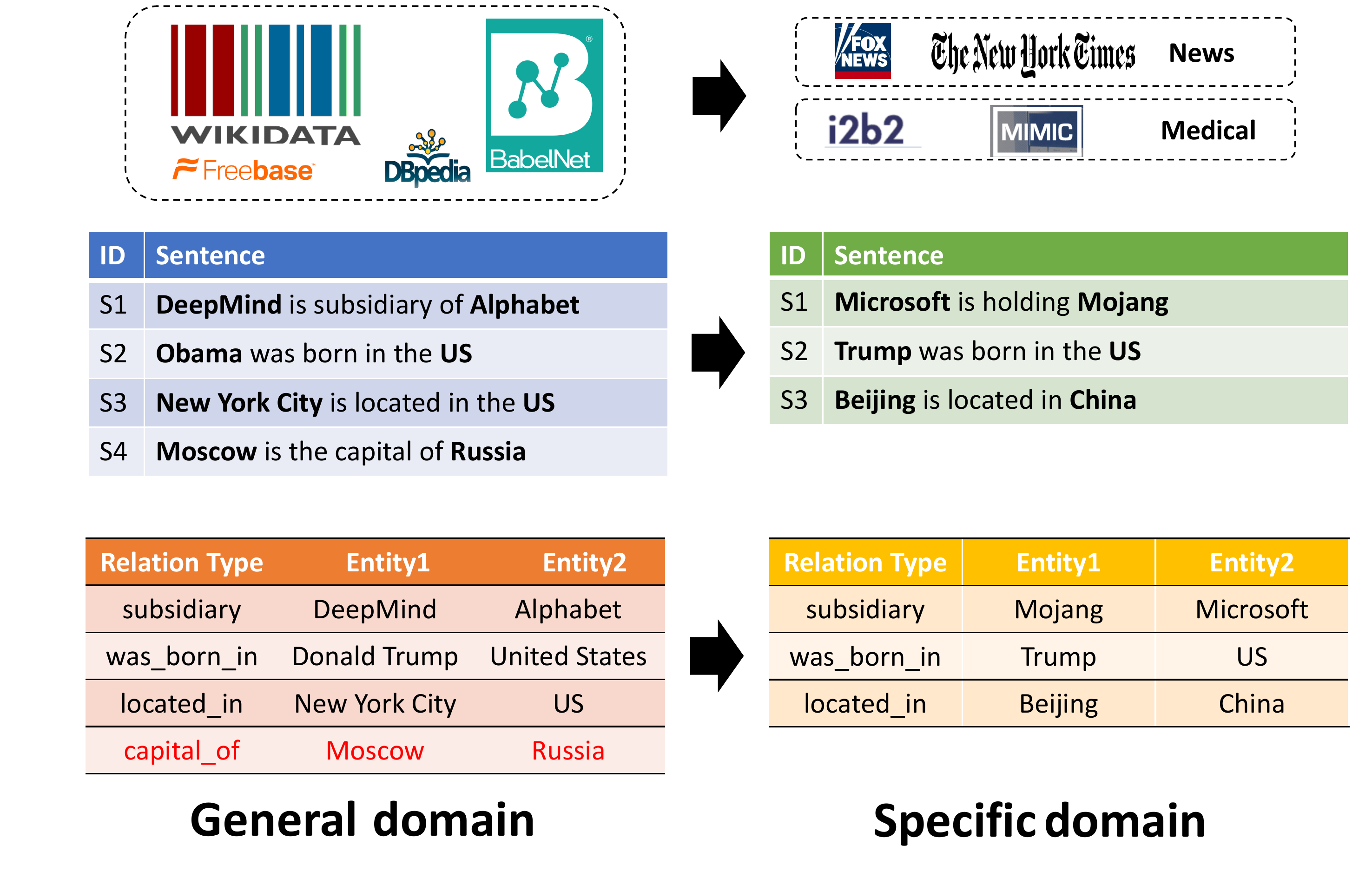}
\caption{Knowlege transfer for RE from the general  domain  (e.g.,Wikipedia)  to  specific  domains.} 

  \label{fig: pic1}
\end{figure}

Domain adaptation (DA) methods \cite{pan2010survey} enable leveraging labeled data from a different but related domain, which is beneficial to RE. On the one hand, it is beneficial to adapt from a fully labeled source domain to a similar but less labeled target domain.  On the other hand,  it is beneficial to adapt from a general domain (e.g., Wikipedia) to a specific domain (e.g., financial,  medical domain).  Moreover, it is beneficial to apply adaptations from a domain with high-quality labels to a domain with noisy labels in the DS settings. However, as shown in Figure \ref{fig: pic1}, there are at least three challenges when adapting a RE system to a new domain short of labels or with noisy labels.

\begin{itemize}
\item  \textbf{Linguistic variation.} First, the same semantic relation can be expressed using different surface patterns in different domains. For example, the relation  \emph{subsidiary} can be expressed such as "DeepMind is a subsidiary of Alphabet" "Microsoft is holding Mojang." It is challenging to learn general domain-invariant textual features that can disentangle the factors of linguistic variations underlying domains and close the linguistic gap between domains.
\item \textbf{Imbalanced relation distribution.} Second, the marginal distribution of relation types varies from domain to domain. For example, a domain about GEO locations may consist of a large number of relational facts about \emph{located\_in}, whereas a domain about persons may be more focused on  \emph{was\_born\_in}. Although the two domains may have the same set of relations, they probably have different marginal distributions on the relations.  This can lead to a \emph{negative transfer} phenomenon \cite{rosenstein2005transfer}  where the out-of-domain samples degrade the performance on the target domain.
\item \textbf{Partial Adaptation.} Third, existing adaptation models generally assume the same label spaces across the source and target domains. However,  it is a common requirement to \emph{partially adapt} from a general domain such as Wikipedia to a small vertical domain such as news or finance that may have a smaller label space. For instance, as shown in Figure \ref{fig: pic1}, by using Wikidata that consists of more than 4,000 relations  as a general source domain and a scientific dataset in a specific domain with a few relations as a target domain, relation type \emph{capital\_of} will trigger the  \emph{negative transfer} problem when discriminating the target relation types \emph{subsidiary} and \emph{was\_born\_in}.
\end{itemize}

To address the aforementioned issues, we propose a general framework called relation-gated   adversarial learning (R-Gated), which consists of three modules: (1) Instance encoder, which learns transferable features that can disentangle the explanatory factors of \textbf{linguistic variations} cross domains. We implement the instance encoder with a convolutional neural network (CNN) considering both model performance and time efficiency. Other neural architectures such as recurrent neural networks (RNN) can also be used as sentence encoders. (2) Adversarial domain adaptation, which looks for a domain discriminator that can distinguish between samples having different relation distributions. Adversarial learning helps learn a neural network that can map a target sample to a feature space such that the discriminator will no longer distinguish it from a source sample. (3) The relation-gate mechanism, which identifies the unrelated source data and down-weight their importance automatically to tackle the problem of negative transfer introduced either  by  \textbf{imbalanced relation distribution} or \textbf{partial transfer}.


\begin{figure*}
\centering
\includegraphics [width=1\textwidth]{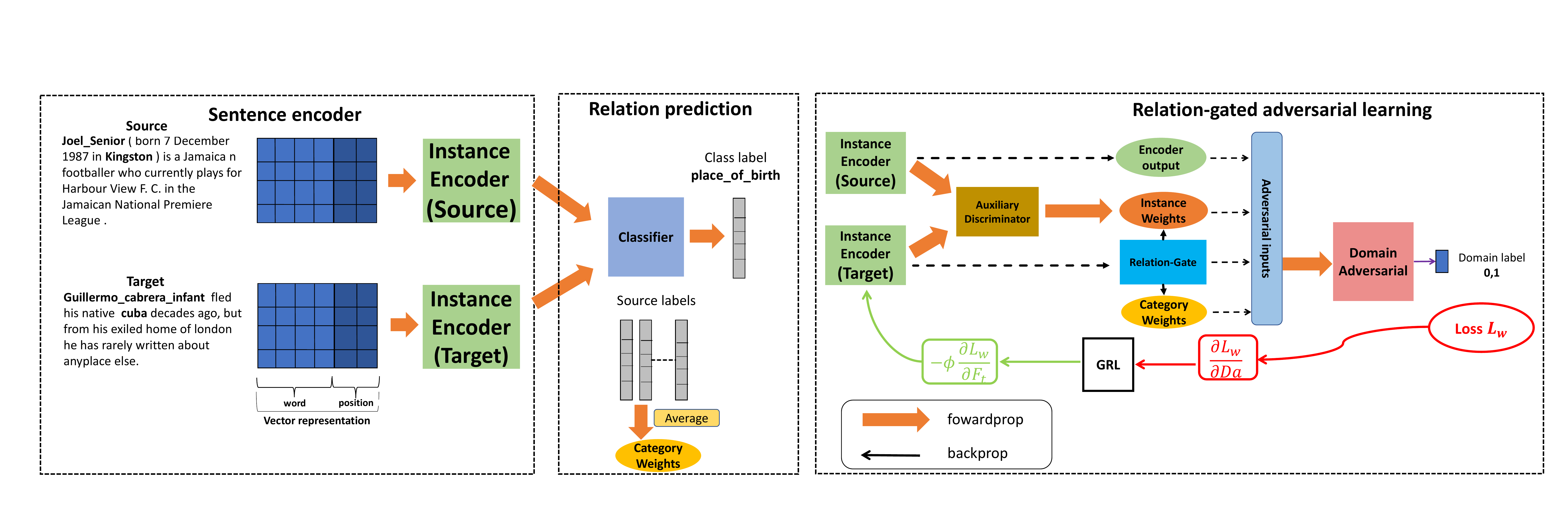}
\caption{Overview of our approach. The parameters of the instance encoder for the source ($F_s$) and relation classifier $C$ are pre-learned and subsequently fixed.  The yellow part denotes the probabilities of assigning the target data to the source classifier to obtain   \textbf{category weights}.   The orange part denotes the output probabilities of auxiliary domain discriminator to obtain   \textbf{instance  weights}. The blue part denotes the \textbf{relation-gate}.  
The red part denotes the traditional adversarial domain classifier. GRL \cite{ganin2016domain} denotes the  gradient reversal layer.} 
\label{fig:architecture}
\end{figure*}

\section{Related Work}
\textbf{Relation Extraction.} RE aims to detect and categorize semantic relations between a pair of entities. To alleviate the annotations given by human experts,
weak supervision and distant supervision have been employed to automatically generate annotations based on KGs (or seed patterns/instances) \cite{zeng2015distant,lin2016neural,ji2017distant,he2018see,zhang2018capsule,zeng2018large,qin2018dsgan,zhang2019long}.   
However, all these models merely focus on extracting facts from a single domain, ignoring the rich information in other domains.

Recently, there have been only a few studies on DA  for RE \cite{plank2013embedding,nguyen2014robust,nguyen2014employing,nguyen2015semantic,fu2017domain}.
 Of these, \cite{nguyen2014robust} followed the supervised DA paradigm.
 In contrast, \cite{plank2013embedding,nguyen2014employing}
  worked on unsupervised DA.  \cite{fu2017domain,rios2018generalizing} presented adversarial learning algorithms for unsupervised DA tasks. However, their methods suffer from the negative transfer bottleneck when encountered partial DA.  To the best of our knowledge, the current approach is the first partial DA work in RE  even in NLP.

\textbf{Adversarial Domain Adaptation.} Generative adversarial nets (GANs) \cite{goodfellow2014generative}  have become a popular solution to reduce domain discrepancy through an adversarial objective concerning a domain classifier \cite{ganin2016domain,tzeng2017adversarial,shen2018wasserstein}.  Recently, only a  few  DA  algorithms \cite{cao2017partial, chen2018re} that can handle  \textbf{imbalanced relation distribution} or  \textbf{partial adaptation} have been proposed.  
\cite{long2018} proposed a method to simultaneously alleviate negative transfer by down-weighting the data of outlier source classes in category level.  \cite{zhang2018importance}
 proposed an   adversarial nets-based partial domain adaptation method to identify the source samples in instance level. 

However,  most of these studies concentrate on image classification. There is a lack of systematic research on adopting DA for NLP tasks. Different from images, the text is more diverse and nosier.  We believe these methods may transfer to the RE setting, but the effect of exact modifications is not apparent. We   make the very first attempt to investigate the empirical results of these methods for RE.  Moreover,  we  propose a relation-gate  mechanism to explicitly model both  coarse-grained   and fine-grained   knowledge transfer   to  lower the negative transfer  effects from  categories and samples. 

\section{Methodology}

\subsection{Problem Definition}
 Given a source domain $D_s = \{(x_i^s,y_i^s\}_{i=1}^{n_s}$ of $n_s$ labeled samples   drawn from distribution $p_s(x)$ associated with  $|C_s|$ classes and a target domain $D_t = \{x_i^t\}_{i=1}^{n_t}$ of $n_t$ unlabeled samples  drawn from distribution $p_t(x)$  associated with $|C_t|$  classes, where $C_t$ is a  subset of $C_s$, we have $|C_s| \ge |C_t|$ in partial DA. We  denote classes  $ c \in C_s$ but  $c \notin  C_t$ as  \emph{outlier classes}.  The goal of this paper is to design a deep neural network that enables learning of transferable features
$f = F_t (x)$ and adaptive classifier $y = C(f)$ for the target domain.

\subsection{Instance Encoder}
Given a sentence $s = \{w_1, ..., w_L\}$, where $w_i$ is the $i$-th word in the sentence, the input is a matrix consisting of $L$ vectors $X = [x_1, ..., x_L]$, where $x_i$ corresponds to $w_i$ and consists of  word embedding and its position embedding \cite{zeng2014relation}. We  apply  non-linear transformation to the vector representation of $X$ to derive a feature vector $f = F(X; \theta)$.  We choose two convolutional neural architectures, CNN \cite{zeng2014relation} and PCNN \cite{zeng2015distant} to encode input embeddings into instance embeddings. Other neural architectures such as RNN \cite{zhang2015relation} and more sophisticated approaches such as ELMo \cite{peters2018deep} or BERT \cite{devlin2018bert} can also be used.

We adopt the unshared feature extractors
for both domains since unshared  extractors are able to capture more domain specific features \cite{tzeng2017adversarial}.  We  train the source discriminative model $C(F_s(x;\theta))$ for the classification task by learning the parameters of the source feature extractor $F_s$ and classifier $C$:

\begin{equation}
\min \limits_{F_s,C} \mathbb{L}_s =\mathbb{E}_{x,y  \sim p_s(x,y)} L(C(F_s(x)),y)
\end{equation} 
where $y$ is the label of the source data $x$,  $L$ is the loss function for classification. Afterwards, the parameters of $F_s$ and $C$ are fixed. Notice that, it is easy to  obtain a  pretrained   RE model from the source domain, which is convenient in real scenarios.

\subsection{Adversarial Domain Adaptation}
To address the issue of  \textbf{linguistic variations} between domains, we utilize adversarial DA, which is a popular solution in both computer vision \cite{tzeng2017adversarial} and  NLP \cite{shah2018adversarial}.
The general idea of adversarial DA is to learn both class discriminative and domain invariant features, where the loss of the label predictor of the source data is minimized while the loss of the domain classifier is maximized. 



\subsection{Relation-Gate  Mechanism}
To address the issue of  \textbf{imbalanced relation distribution} and \textbf{partial adaptation}, we introduce a relation-gate mechanism to explicitly  model instance and category impact in the source domain.  

\textbf{Category  Weights Learning.}
Given that not all classes in the source domain are beneficial and can be adapted to the target domain,   it is intuitive to assign different weights to different classes to lower the negative transfer effect of outlier classes in the source domain as the target label space is a subset of the source label space.   For example,  given that the relation \emph{capital\_of} in the source domain does not exist in the target domain as shown in Figure \ref{fig: pic1}, it is necessary to lower this relation to mitigate negative transfer. 


We average the label predictions on all target data from the source classifier as class weights \cite{long2018}.  Practically,  the source classifier $C(F_s(x_i))$  reveals a probability distribution over the source label space $C_s$. This distribution  characterizes well the probability of assigning $x_i$ to each of the $|C_s|$ classes.   We  average the label predictions $\hat y_i = C(F_s(x_i)),\ x_i \in D_t , $ on all target data since it is possible that the source classifier can make a few mistakes on some target data and assign large probabilities to false classes or even to outlier classes. The weights indicating the contribution of each source class to the training can be computed as follows:
\begin{equation}
w^{category} = \frac{1}{n_t}\sum_{i=1}^{n_t} \hat y_i \label{category}
\end{equation}
where $w^{category}$ is a $|C_s|$-dimensional weight vector quantifying the contribution of each source class.  

\textbf{Instance  Weights Learning.} Although the category  weights provide a global weights mechanism to de-emphasize the effect of outlier classes, different instances have different impacts, and not all instances are transferable. Considering the relation \emph{educated\_at} as an example,  given an instance \emph{"\textbf{James Alty} graduated from \textbf{Liverpool University}"} from target domain, semantically, a   more similar  instance of \emph{"\textbf{Chris Bohjalian} graduated from \textbf{Amherst College}"}  will  provide more  reference  while  a dissimilar  instance \emph{"He was a professor at \textbf{Reed College} where he taught \textbf{Steve Jobs}"} may have little contributions. It is necessary to learn fine-grained instance  weights to lower the effects of samples that are nontransferable.

Given the sentence encoder of  the source and target domains, we utilize a pretrained auxiliary domain classifier for instance  weights learning. We regard the output of the optimal parameters of the auxiliary domain classifier as instance weights. The concept is that if the activation of the auxiliary domain classifier is large, the sample can be almost correctly discriminated from the target domain by the discriminator, which means that the sample is likely to be nontransferable  \cite{zhang2018importance}.

Practically, given the learned $F_s(x)$ from the  instance encoder, a domain adversarial loss is used to reduce the shift between domains by optimizing $F_t(x)$ and auxiliary domain classifier $D$:
\begin{equation}\begin{split}
\min \limits_{F_t} \max \limits_{D} \mathbb{L}_d (D,F_s,F_t) =& \mathbb{E}_{x \sim p_s(x)}[logD(F_s(x))]+ \\&
\mathbb{E}_{x \sim p_t(x)}[1-logD(F_t(x))]
\label{eq: firstgan}
\end{split}
\end{equation}
To avoid a degenerate solution, we initialize $F_t$ using the parameter of $F_s$. The auxiliary domain classifier is given by $D(f) = p(y=1|x)$ where x is the input from the source and the target domains.    If  $D(f) \equiv 1$, then it is  likely that the sample is nontransferable, because it  can be almost perfectly discriminated from the target distribution by the domain classifier. The contribution of these samples should be small. Hence, the weight function should be inversely related to $D(f)$, and a natural way to define the  weights of the source samples is:

\begin{equation}
w_i^{instance} = \frac{1}{\frac{D(F_s(x))}{D(F_t(x))}+1} = 1-D(f)
\end{equation}

\textbf{Relation-Gate.} 
Both  category and instance  weights are helpful.  However, it is obvious that the weights of different  granularity  have different contributions to different target  relations.  On the one hand, for  target relations (e.g., \emph{located\_in}) with relatively  less semantically similar  source relations,  it is advantageous to strengthen the category weights  to reduce the  negative effects of outlier classes. On the other hand, for target relations (e.g., \emph{educated\_in}) with   many semantically similar  source relations (e.g., \emph{live\_in},   \emph{was\_born\_in}),  it is difficult to differentiate the impact of different source relations, which indicates the necessity of learning fine-grained instance weights.

For an instance in the source domain with label $y_j$, the weight of this instance is:
\begin{equation}
w_i^{total}=\alpha w_i^{instance}+(1-\alpha) w_{j}^{category}
\end{equation}
where  $w_{j}^{category}$ is the value in the $j$th-dimension of $w^{category}$.  We normalize the weight $w_i^{total} = \frac{n_sw_i^{total}}{\sum_{i=1}^{n_s}w_i^{total}}$.
$\alpha$  is the output  of relation-gate  to explicitly balance the  instance and category weights which is  computed as below.
\begin{equation}
\alpha = \sigma(W_rF_t(x))
\end{equation}
where $\sigma$ is the activation function, $W_r$ is the weight matrix. 
\subsection{Initialization and Implementation Details}

The overall objectives of  our approach  are  $\mathbb{L}_s$, $\mathbb{L}_d$ and:
\begin{equation}
\begin{split}
\min \limits_{F_t} \max \limits_{D_a} &  \mathbb{L}_w(C,D_a,F_s,F_t)=\\
& \mathbb{E}_{x \sim p_s(x)}[w^{total}logD_a(F_s(x))]+  \\
&   \mathbb{E}_{x \sim p_t(x)}[1-logD_a(F_t(x))]
\end{split}\label{eq: allpro}
\end{equation}
 where  $D_a$ is the domain adversarial. Note that,  weights $w^{total}$\footnote{The weights can be updated in an iterative fashion when $F_t$ changes. However, we found no improvement in experiments, so we compute the weights and  fix them.} are automatically computed and assigned to the source domain data to de-emphasize the outlier classes and nontransferable instances regarding partial DA, which can mitigate  negative transfer. The overall training procedure\footnote{Training details and  hyper-parameters  settings can be found in  supplementary materials} is shown below.


\begin{algorithm}[th]
\caption{Overall Training Procedure}
\label{algorithm: training}
1.Pre-train $F_s$ and $C$  on the source domain and fix  all parameters afterward.

2.Compute category weights by Equation \ref{category}.

3.Pre-train  $F_t$ and $D$ by  Equation \ref{eq: firstgan} and compute instance  weights, then fix parameters of $D$.

4.Train $F_t$ and $D_a$ by Equation \ref{eq: allpro}, update the parameters  of $F_t$  through GRL.
\end{algorithm}

\section{Experiments}

\subsection{Datasets and  Evaluation}

\textbf{ACE05 Dataset.}
We use the ACE05\footnote{https://catalog.ldc.upenn.edu/LDC2006T06} dataset to evaluate our approach by dividing the articles from its six genres into respective domains: broadcast conversation (bc), broadcast news (bn), telephone conversation (cts), newswire (nw), usenet (un) and weblogs (wl).  We use the same data split followed by \cite{fu2017domain}, in which bn $\&$ nw are used as the source domain, half of bc, cts, and wl are used as the target domain for training (no label available in the unsupervised setting), and  the other half of bc, cts, and wl  are used as  target domain for test. We  split  10\% of the training set to form the development set to fine-tune hyper-parameters such as $\alpha$.  We conducted two kinds of experiments. The first is normal DA, in which the source and target domain have the same classes. The second is partial DA, in which the target domain has only half of the source domain classes.

\textbf{Wiki-NYT Dataset.}
 For DS setting, we utilize two existing datasets NYT-Wikidata \cite{zeng2016incorporating},  which align Wikidata with New York Times corpus (NYT), and  Wikipedia-Wikidata \cite{sorokin2017context},  which align Wikidata with   Wikipedia. We filter 60 shared relations to construct a new dataset  Wiki-NYT\footnote{We will release our dataset.}, in which Wikipedia is the source domain and NYT corpus is the target domain.  We split the dataset into three sets: 80\% training, 10\% dev, and 10\% test. We conducted partial DA experiments (60 classes  $\rightarrow$ 30 classes). We randomly choose half of the classes to sample the target domain data.

\subsection{Parameter Settings}
To fairly compare the results of our models with those baselines, we set most of the experimental parameters following \cite{fu2017domain,lin2016neural}. 
We train GloVe \cite{pennington2014glove} word embeddings on the Wikipedia and  NYT corpus with 300 dimensions. In both the training and test set, we truncate sentences with more than 120 words into 120 words. 

\subsection{Evaluation Results on ACE05}
To evaluate the performance of our proposed  approach, we compared our model with various DA  models: \textbf{CNN+R-Gated} is our approach,  \textbf{CNN+DANN} is an unsupervised adversarial DA method \cite{fu2017domain},  \textbf{Hybrid} is a composition model that combines  traditional feature-based method, CNN and RNN \cite{nguyen2015combining}, and \textbf{FCM} is a compositional embedding model.  From the evaluation results as shown in Table \ref{normalace}, we observe that (1) our model achieves performance comparable to that of CNN+DANN, which is a state-of-the-art model, in normal DA scenario and significantly outperforms the vanilla models without adversarial learning. This shows that domain adversarial learning is effective for learning domain-invariant features to boost performance. (2)  Our model significantly outperforms the plain adversarial DA model, CNN+DANN, in partial DA. This demonstrates the efficacy of our hybrid weights mechanism\footnote{Since adversarial DA method significantly outperforms traditional methods \cite{fu2017domain}, we skip the performance comparison with FCM and Hybrid for partial DA.}.

\begin{table}[!htbp]
\centering
\begin{small}
\begin{tabular}{c|c|c|c|c}

\hline

\centering \textbf{Normal DA} & \centering \textbf{bc}& \centering \textbf{wl}& \centering \textbf{cts}&   \textbf{avg}\\

\hline

\centering FCM&61.90 & N/A& N/A& N/A  \\

\centering Hybrid&63.26&N/A&N/A &N/A\\

\centering CNN+DANN &65.16&55.55&  57.19 &59.30\\
\hline

\centering CNN+R-Gated &\textbf{66.15*}&\textbf{56.56*}&56.10& \textbf{59.60*}\\
\hline
\hline
\centering \textbf{Partial DA} & \centering \textbf{bc}& \centering \textbf{wl}& \centering \textbf{cts}&   \textbf{avg}\\
\hline
\centering CNN+DANN&63.17&53.55&  53.32 &56.68\\
\hline
\centering CNN+R-Gated &\textbf{65.32*}&\textbf{55.53*}&\textbf{54.52*}&\textbf{58.92*} \\
\hline
\end{tabular}
\caption{F1 score of normal and partial DA on ACE05 dataset. * indicates $p_{value} < 0.01$ for t-test evaluation. }\label{normalace}
\end{small}
\end{table}

\subsection{Evaluation Results  on Wiki-NYT}
For DS setting, we consider the setting of (1) \textbf{unsupervised adaptation} in which the target labels are removed, (2) \textbf{supervised adaptation} in which the target labels are retained to fine-tune our model.

\begin{figure}[H]
  \centering
  \includegraphics[width=0.38\textwidth]{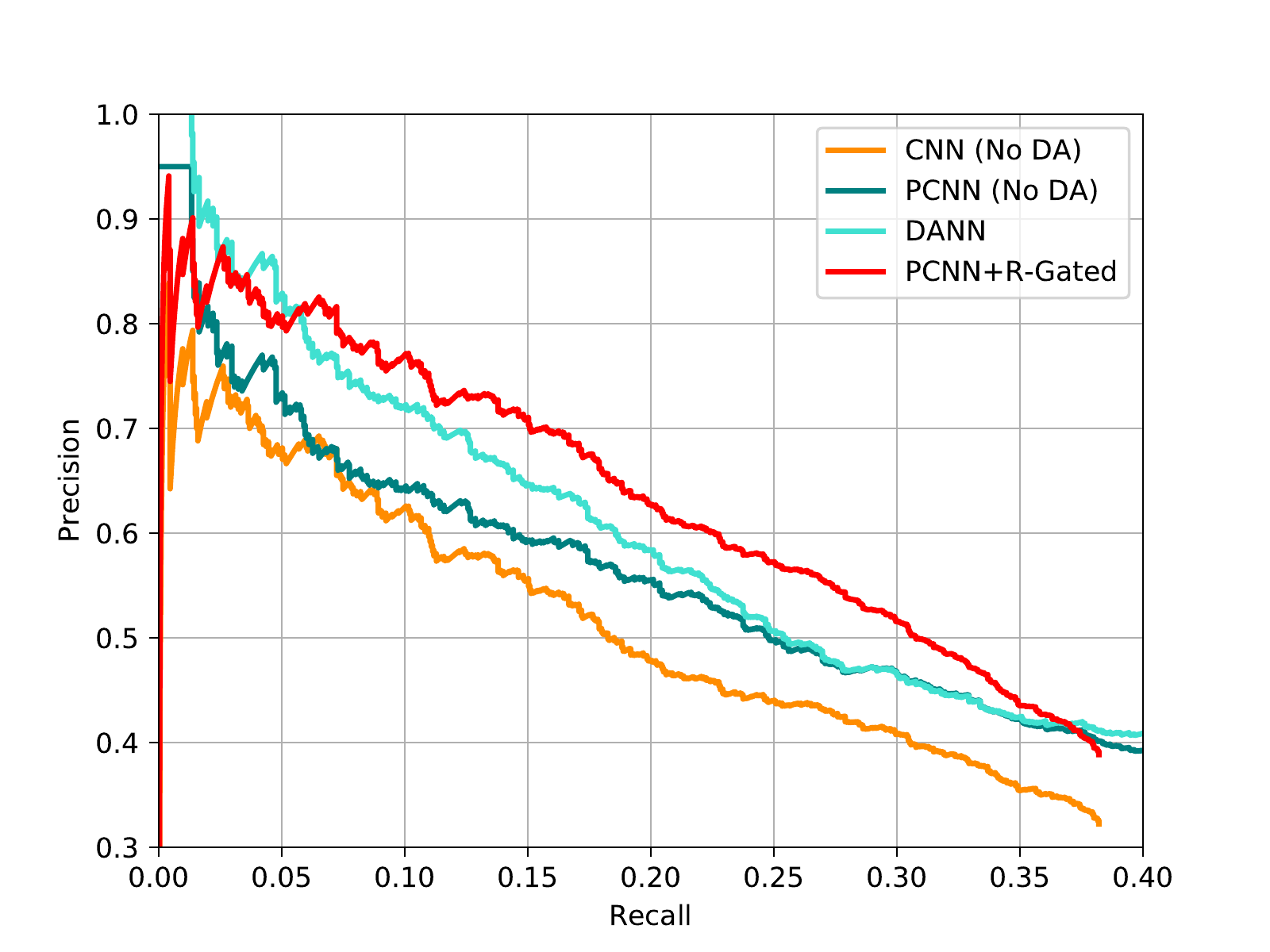}
  \caption{Unsupervised adaptation results.}\label{fig: wikinytpr}
\end{figure}

\textbf{Unsupervised Adaptation.} Target labels are unnecessary in unsupervised Adaptation. We report the results  of our approach and various  baselines: \textbf{PCNN+R-Gated} is our unsupervised  adaptation approach, \textbf{PCNN (No DA)}  and \textbf{CNN (No DA)} are the methods trained on the source domain by  PCNN \cite{lin2016neural} and CNN  \cite{zeng2014relation} and tested on the target domain. Following  \cite{lin2016neural}, we perform both held-out evaluation as the precision-recall curves shown in Figure \ref{fig: wikinytpr} and manual evaluation in which we manually check the top-500 prediction results, as shown in Table \ref{unsupervise}. 

We observe that (1) our   approach    achieves  the best performance among all the other unsupervised DA models, including  CNN+DANN. This further demonstrates the effectiveness of hybrid weights mechanism. (2) Our unsupervised DA model achieves nearly the same performance even with the supervised approach CNN; however, it does not outperform PCNN. This setting could be advantageous as in many practical applications,  the knowledge bases in a vertical target domain may not exist at all or must be built from scratch.
\begin{figure}[H]
  \centering
  \includegraphics[width=0.38\textwidth]{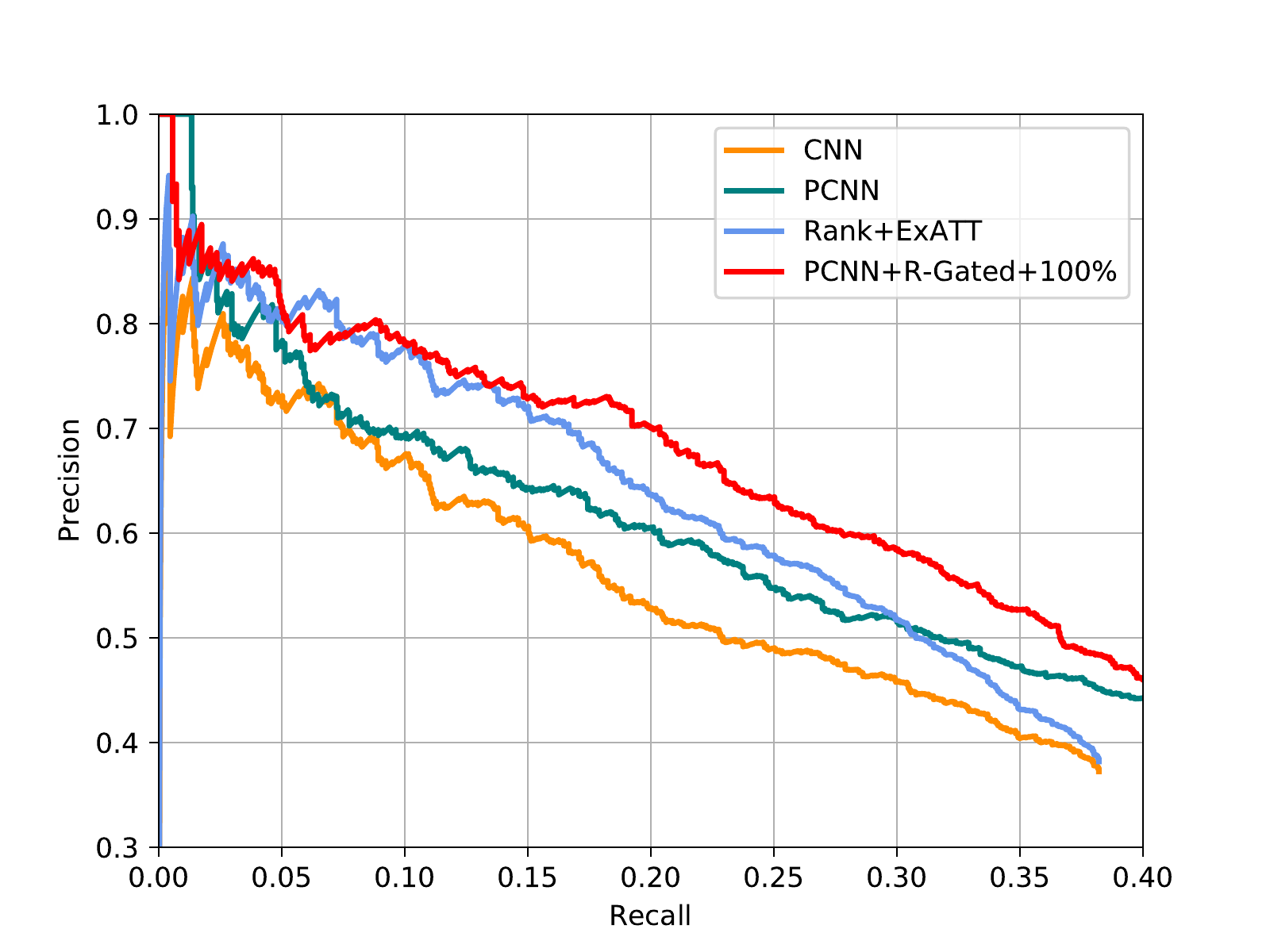}
  \caption{Supervised adaptation results.}\label{fig:pretrain}

\end{figure}
\textbf{Supervised Adaptation.} Supervised Adaptation does require labeled target data; however, the target labels might be few or noisy. In this setting, we fine-tune our model with target labels. We report the results of our approach and various baselines: \textbf{PCNN+R-Gated+$k\%$} implies fine-tuning our model using $k\%$ of the target domain data, \textbf{PCNN} and \textbf{CNN} are the methods trained on the target domain by PCNN \cite{lin2016neural} and CNN \cite{zeng2014relation}, and \textbf{Rank+ExATT}  is the method trained on the target domain which  integrates PCNN with a pairwise ranking framework \cite{ye2017jointly}.

As shown in  Figure \ref{fig:pretrain} and Table \ref{unsupervise}, we observe that (1) our fine-tuned model +100\% outperforms both CNN and PCNN and achieves results comparable to that of Rank+ExATT. The case study results in Table \ref{case3} further shows that our model can correct noisy labels to some extent due to the relatively high quality of source domain data. (2) The extent of improvement from using 0\% to 25\% of target training data is consistently more significant than others such as using 25\% to 50\%, and fine-tuned model with only thousands labeled samples (+25\%) matches the performance of training from scratch with 10$\times$  more data, clearly demonstrating the benefit of our approach. (3) The top 100 precision of fine-tuned model degrades from 75\% to 100\%. This indicates that there exits noisy data which contradict with the data from the source domain. We will address this by adopting  additional denoising mechanisms  like  reinforcement learning,  which  will be part of our future work. 

\begin{table}[!htb]    
\renewcommand\tabcolsep{4pt}
\centering

\begin{small}
\begin{tabular}{c|cccc}

\hline

\centering Precision & Top 100 &Top 200&Top 500&Avg. \\

\hline

\centering CNN (No DA)&0.62& 0.60  & 0.59&0.60\\
\centering PCNN (No DA)&0.66& 0.63  & 0.61&0.63\\

\centering CNN+DANN&0.80& 0.75& 0.67&0.74\\
\hline

\centering CNN & 0.85& 0.80& 0.69& 0.78\\
\centering PCNN&  0.87& 0.84& 0.74& 0.81 \\
\centering Rank+ExATT&  0.89& 0.84& 0.73& 0.82 \\
\hline
\centering PCNN+R-Gated&\textbf{0.85*} &\textbf{0.83*}  & \textbf{0.73*}& \textbf{0.80*}\\
\centering +25\% &0.88 &0.84 & 0.75& 0.82\\
\centering +50\% &0.89 &0.85  & 0.76& 0.82\\
\centering +75\% &0.90 &0.85  & 0.77& 0.83\\
\centering +100\% &0.88 &\textbf{0.86*}  & \textbf{0.77*}& \textbf{0.83*}\\

\hline
\end{tabular}
\caption{Precision values of the top 100, 200 and 500 sentences for  unsupervised and supervised adaptation. * indicates $p_{value} < 0.01$ for t-test evaluation.
}\label{unsupervise}
\end{small}
\end{table}

\subsection{Ablation Study} To better demonstrate the  performance of  different strategies in our model, we  separately remove the category and  instance weights.  The  experimental results on Wiki-NYT dataset are summarized in Table \ref{tab:ablation}. \textbf{PCNN+R-Gated} is our method;  \textbf{w/o gate}  is the method  without relation-gate ($\alpha$  is fixed.); \textbf{w/o category}  is the method  without category   weights  ($\alpha=1$) ;   \textbf{w/o instance} is the method  without instance  weights ($\alpha=0$) ;  \textbf{w/o both} is the method without both weights. 
We observe that  (1) the performance significantly degrades when we remove "relation-gate." This is reasonable because the category and instance play different  roles for different relations, while \textbf{w/o gate}  treat weights equally which hurts the performance. (2) the performance   degrades when we remove "category weights"  or "instance weights." This is reasonable because  different weights have  different  effects   in de-emphasizing those outlier classes or instances.

\begin{table}[!htb]    
\renewcommand\tabcolsep{4pt}
\centering

\begin{small}
\begin{tabular}{c|cccc}

\hline

\centering Precision & Top 100 &Top 200&Top 500&Ave. \\

\hline

\centering PCNN+R-Gated&\textbf{0.85*} &\textbf{0.83*}  & \textbf{0.73*}& \textbf{0.80*}\\ \hline
\centering w/o  gate&0.85 &0.79  &0.70&0.78\\
\centering w/o category &0.81 &0.76  & 0.66& 0.74\\
\centering w/o instance &0.84 &0.78  & 0.69& 0.77\\
\centering w/o  both &0.80& 0.75& 0.65&0.73\\

\hline
\end{tabular}
\caption{Precision values of the top 100, 200 and 500 sentences for  ablation study.  * indicates $p_{value} < 0.01$ for t-test evaluation.}\label{tab:ablation}
\end{small}
\end{table}

\subsection{Parameter Analysis}
\begin{figure} \centering
\subfigure[$\alpha$ w.r.t \#Relations] { 
  \includegraphics[width=0.22\textwidth]{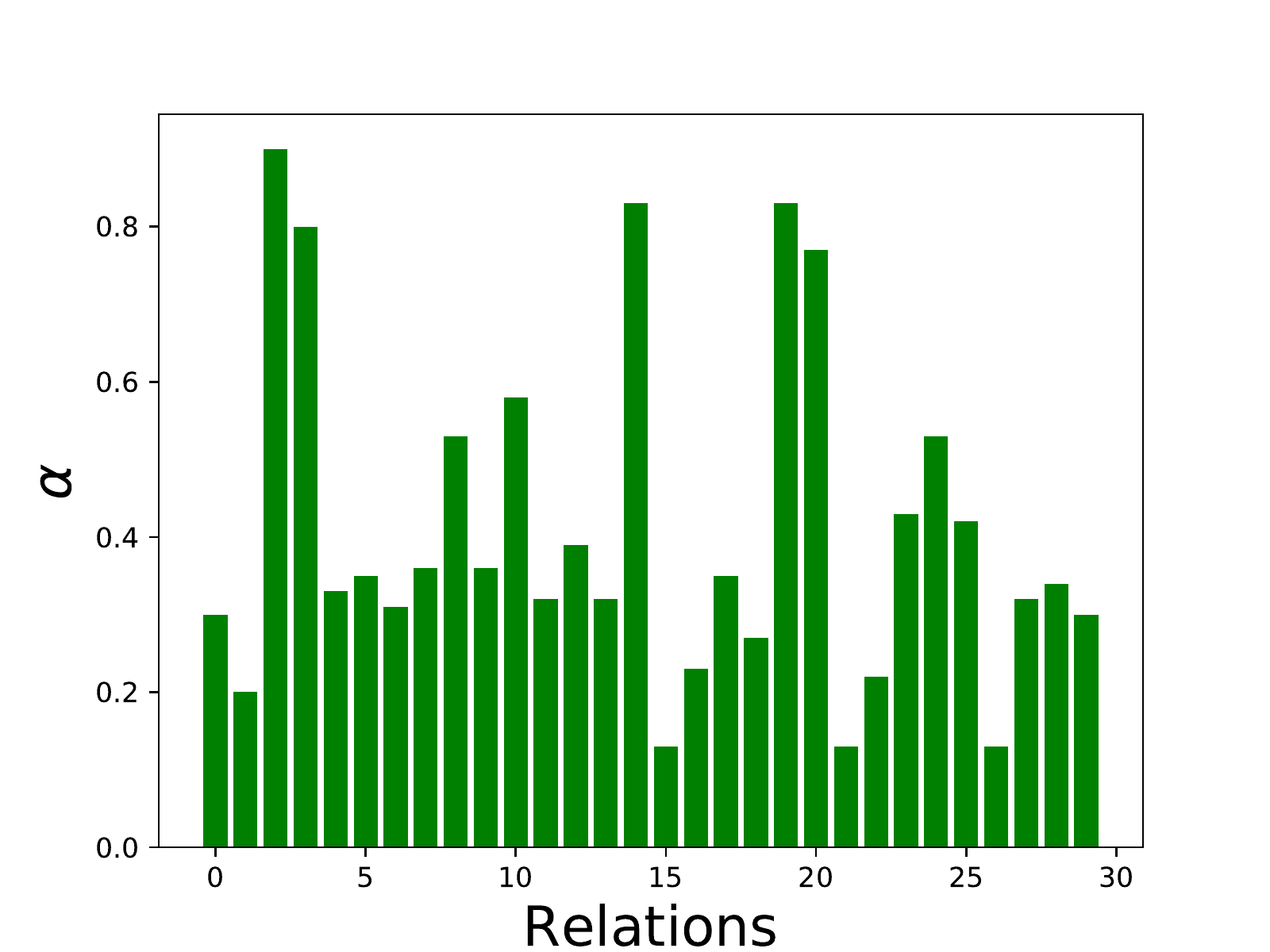}
}
\subfigure[F1 w.r.t \#Target Classes] { 
\includegraphics[width=0.22\textwidth]{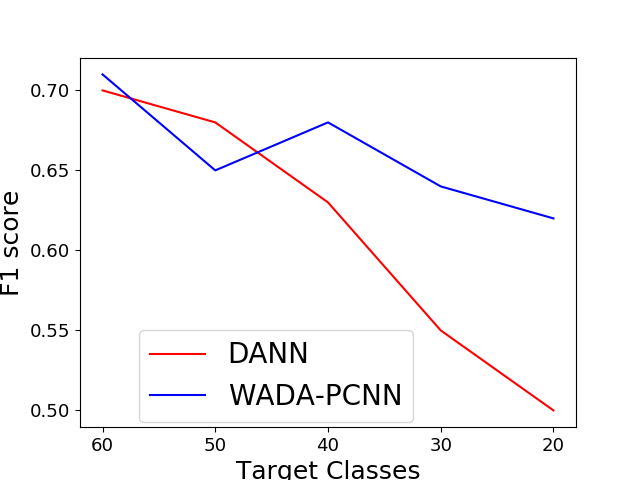}
}
\caption{Parameter analysis  reuslts.}
\label{fig}
\end{figure}

\textbf{Relation-Gate.} To   further explore the effects of relation-gate, we visualize $\alpha$ for all  target relations on Wiki-NYT dataset.  From the results shown in Figure \ref{fig} (a), we observe the following: (1)
the instance and category weights have different influences on performance for different relations.  Our relation-gate  mechanism   is powerful to find that   instance weights is more important for those relation (e.g., \emph{educated\_at}, \emph{live\_in}    
a.k.a.,  relations with highest $\alpha$  score in  Figure \ref{fig} (a)) while category  weights are more useful for other relations.  (2) The category weights have relatively more  influence  on the performance than instance weights  for most  of the relations due to the noise and variations in instances; however, the category weights are averaged on  all  target  data and thus less noisy. 
 
\textbf{Different Number of Target Classes.} We investigated  partial DA by varying the number of target classes in the Wiki-NYT dataset\footnote{The target classes are sampled three  times randomly, and the results are averaged.}. Figure \ref{fig} (b) shows that when the number of target classes decreases, the performance of CNN+DANN degrades quickly, implying the severe  negative transfer. We observe that  PCNN+R-Gated   outperforms CNN+DANN when the number of target classes decreases. Note that, PCNN+R-Gated performs comparably to CNN+DANN in standard DA when the number of target classes is 60. This means that the weights mechanism will not wrongly filter out classes when there are no outlier classes.

\subsection{Case Study}
We select  samples  from shared and outlier relations for detail analysis in \textbf{Case 1}, \textbf{Case 2} and \textbf{Case 3} and    give examples to show that our model can correct  noisy labels in \textbf{Case 4}. 

\textbf{Case 1: Relation-gate.}
We give some examples of how our relation-gate balance the weights for classes and instances. In Table \ref{case12}, we display the $\alpha$ of different relations. For  relation \emph{capital\_of}, there are lots of dissimilar relations so  category  weights are more important, which results in a small  $\alpha$. For  relation \emph{educated\_in} (\emph{edu\_in}), the instance difference is more important so $\alpha$  is relatively large.

\textbf{Case 2: Category Weights.}
We give some examples of how our approach assign different weights for classes to mitigate the negative effect of outlier classes. In Table \ref{case12}, we display the sentences  from shared classes and outlier classes. The relation \emph{capital\_of}  is an outlier class whereas \emph{director} is a shared class. We observe that our model can automatically find outlier classes and assign lower weights to them.

\begin{table}[h]\centering
    \fontsize{8.5}{10}\selectfont
    \begin{tabular}{|p{3.7cm}|p{0.9cm}|p{0.3cm}|p{0.3cm}|p{0.3cm}|}
    \hline 
    \centering Instances&Relations&$w^{c}$&$w^{i}$& $\alpha$\\
     \hline
    He was born in \textbf{Rio\_de\_Janeiro}, \textbf{Brazil} to a German father and a Panama nian mother.
&capital\_of&0.1&0.2&0.1   \\
   \hline
In 2014, he made his Tamil film debut in \textbf{Malini\_22\_Palayamkottai} directed by \textbf{Sripriya}.
 &director&0.7&0.8 &0.4  \\
   \hline
  Sandrich was replaced by \textbf{George\_Stevens} for the teams 1935 film \textbf{The\_Nitwits}.
 &director&0.7&0.5 &0.5 \\
   \hline
     \textbf{Camp} is a 2003 independent musical\_film written and directed by \textbf{Todd\_Graff}.

 &director&0.7&0.3& 0.4  \\ \hline
 \textbf{Chris Bohjalian} graduated from \textbf{Amherst College}&edu\_in&0.4&0.7&0.9\\

   \hline
    \end{tabular}
    \caption{Examples for Case 1, 2 and 3, $w^{c}$ and $w^{i}$ denote category and instance weights, respectively.}
    \label{case12}
\end{table}

\textbf{Case 3: Instance Weights.}
We give some examples of how our approach assign different weights for instances to de-emphasize the nontransferable samples. In  Table \ref{case12}, we observe that (1) our model can automatically assign lower weights to instances in outlier classes. (2)  Our model can assign different weights for instances in the same class space to down-weight the negative effect of nontransferable instances.  (3) Although our model can identify some nontransferable instances, it still assigns incorrect weights to some instances (The end row in Table \ref{case12}) which is semantically similar and transferable. We will address  this by adopting additional mechanisms like transferable attention \cite{wang2019transferable}, which will be part of our future work.

\begin{table}[h]\centering
    \fontsize{8.3}{10}\selectfont
    \begin{tabular}{|p{4.2cm}|p{1cm}|p{1.1cm}|}
    \hline 
   \centering \textbf{Instances}&\textbf{DS}&\textbf{R-Gated} \\
    \hline
They are trying to create a united front at home in the face of the pressures \textbf{Syria} is facing,“ said \textbf{Sami Moubayed}, a political analyst and writer here.&p\_of\_birth& NA\\
\hline
Iran injected \textbf{Syria} with much confidence: stand up, show defiance,“ said \textbf{Sami Moubayed}, a political analyst and writer in Damascus.&p\_of\_birth& NA\\
\hline
   
    \end{tabular}
    \caption{Examples for Case 3, $p\_of\_birth$ is $place\_of\_birth$ for short.}
    \label{case3}
\end{table}

\textbf{Case 4: Noise Reduction.}
We give some examples of how our approach takes effect in correcting the noisy labels of the target domain.  In Table \ref{case3}, we display the sentences that are  wrongly marked in DS settings and show their labels predicted by our approach. We observe that our model can  correct some noisy labels, verifying that our model can be used to adapt from a source domain with high-quality labels to a target domain with noisy distant labels. This is reasonable  because  Wikidata is partly aligned with the NYT corpus, entity pairs with fewer sentences are more likely to be false positive, which is the major noise factor. However,  Wikidata can be relatively better aligned with Wikipedia, which can create more true positive samples.


\section{Conclusion and Future Work}

In this paper, we propose a  novel  model of  relation-gated  adversarial  learning  for RE.  Extensive experiments demonstrate that our model achieves results that are comparable with that of state-of-the-art DA baselines and can improve  the accuracy of distance supervised RE through fine-tuning. In the future, we intend to improve the DA using only a few supervisions, namely few-shot adversarial DA.  It will also be promising to apply our method to other NLP scenarios.


\bibliography{emnlp-ijcnlp-2019}
\bibliographystyle{acl_natbib}

\end{document}